\pdfoutput=1

\documentclass[11pt]{article}

\usepackage[preprint]{acl}
\usepackage{mdframed}
\usepackage{times}
\usepackage{latexsym}
\usepackage{tikz}
\usetikzlibrary{positioning}
\usepackage{enumitem}
\usepackage{booktabs, cellspace}

\usepackage[T1]{fontenc}

\usepackage[utf8]{inputenc}
\usepackage{listings}
\usepackage{microtype}

\usepackage{inconsolata}
\usepackage{graphicx}
\usepackage{balance}

\graphicspath{{images/}}

\title{A Scalable Tool for Measuring Manner and Result Verbs in Developmental Language Research}

\author{
Divyesh Pratap Singh \\
University at Buffalo
\And
Dakshesh Gusain \\
Nanyang Technological University
\And
Federica Bulgarelli \\
University at Buffalo
\AND
Alison Eisel Hendricks \\
University at Buffalo
\And
John Beavers \\
The University of Texas at Austin
\And
Nathan M. Beers \\
University at Buffalo
\AND
Ifeoma Nwogu \\
University at Buffalo
}
\begin{document}
\maketitle

\begin{abstract}
Manner and result verbs encode different aspects of event structure and have been discussed in developmental work as a potentially informative distinction for studying early verb learning. However, this distinction remains difficult to measure at scale because large annotated resources for manner and result classification are not currently available. We present a computational approach for identifying manner and result verbs in sentence context. Using linguistically informed prompts, we generate sentence-level annotations with large language models over data drawn from MASC and InterCorp, extending coverage from previously annotated portions of VerbNet to 436 classes. We then train a RoBERTa-based classifier on these annotations and evaluate it on three held-out gold-standard datasets, including previously annotated items and a new expert-annotated set. Across these evaluations, the model shows promising performance, with average accuracy up to 89.6\%. We present this work as a scalable measurement tool that can support future research on verb semantics in developmental and other language datasets, while noting that further validation is needed for borderline cases, mixed manner/result verbs, and downstream developmental applications.
\end{abstract}

\section{Introduction}\label{sec:intro}
Early language development depends not only on how much language children hear, but also on the kinds of meanings encoded in the words they learn. Verbs are especially important because they support children’s transition to multiword speech and later grammatical development. Verb vocabulary around age two predicts later grammatical outcomes and, for some developmental questions, may be more informative than noun vocabulary \cite{hadley2016toddlers}. These issues are especially relevant for late talkers, whose early language trajectories are heterogeneous and whose later outcomes are difficult to predict from broad lexical measures alone.

One semantic distinction that has become relevant in this literature is the contrast between \emph{manner} and \emph{result} verbs. Manner verbs encode how an action is carried out (e.g., \emph{rub}, \emph{scribble}, \emph{run}), whereas result verbs encode an outcome or change of state (e.g., \emph{clean}, \emph{fill}, \emph{open}) \cite{hovav2010reflections,levin2008constraint}. Developmental work suggests that this distinction may be informative for understanding variation in early verb learning. For example, \citet{horvath2022difference} report that the relative proportions of manner and result verbs differ between late talkers and typically developing children, and that children who produce more manner verbs also tend to produce more verbs overall \cite{horvath2019syntactic,horvath2022difference}. Because a substantial proportion of children with early language delay later meet the criteria for Developmental Language Disorder (DLD), the task of identifying finer-grained semantic properties of children’s early vocabularies may help clarify which aspects of early language are associated with these later outcomes.

At the same time, this distinction remains difficult to study at scale. Although computational linguistics has made substantial progress on grammatical annotation tasks such as part-of-speech tagging \cite{derose1988grammatical}, fine-grained semantic categorization is generally more challenging. Prior work on related event-semantic distinctions suggests that verb meaning is difficult to classify automatically in context \cite{friedrich2022kind,friedrich2017classification,metheniti2022time,friedrich2016situation}. As a result, theoretically important contrasts such as manner versus result verbs still lack broad, scalable annotation resources, limiting their use in developmental language research.

To address this gap, we present a computational approach for identifying manner and result verbs in context. We use large language models (LLMs) as informed annotators, drawing on established linguistic definitions of manner and result verbs together with a small set of illustrative examples. We prompt LLMs to label sentences from the Manually Annotated Sub-Corpus (MASC; \citealt{ide2008masc}) and the InterCorp parallel corpus (\citealt{ek2012case}), expanding coverage from 151 previously annotated VerbNet classes to 436 classes \citep{brown2019verbnet,kipper2008large}. We then fine-tune a pretrained RoBERTa classifier \citep{Liu2019RoBERTaAR} on these labels and evaluate it on three held-out gold-standard datasets. We position this system as a scalable measurement tool that can support research on verb semantics in larger corpora, including developmental language data.

\noindent In summary, our contributions are:
\vspace*{-1em}
\begin{itemize}[leftmargin=*]\setlength\itemsep{0em}
    \item We present a scalable computational framework for identifying \emph{manner} and \emph{result} verbs in sentence context, enabling this theoretically important distinction to be measured in larger language datasets.
    \item We introduce an annotation pipeline that leverages large language models to generate training data for this task in the absence of large-scale gold-standard resources.
    \item We show that a RoBERTa-based classifier trained on these annotations can reliably distinguish manner and result verbs in context.
    \item We will publicly release our code and annotated dataset, extending coverage to 436 VerbNet classes, to support future research.
\end{itemize}

\section{Understanding Verb Root Meaning}\label{sec:meaning}
\begin{figure}[h]
	\centering
	\includegraphics[width=1.0\linewidth]{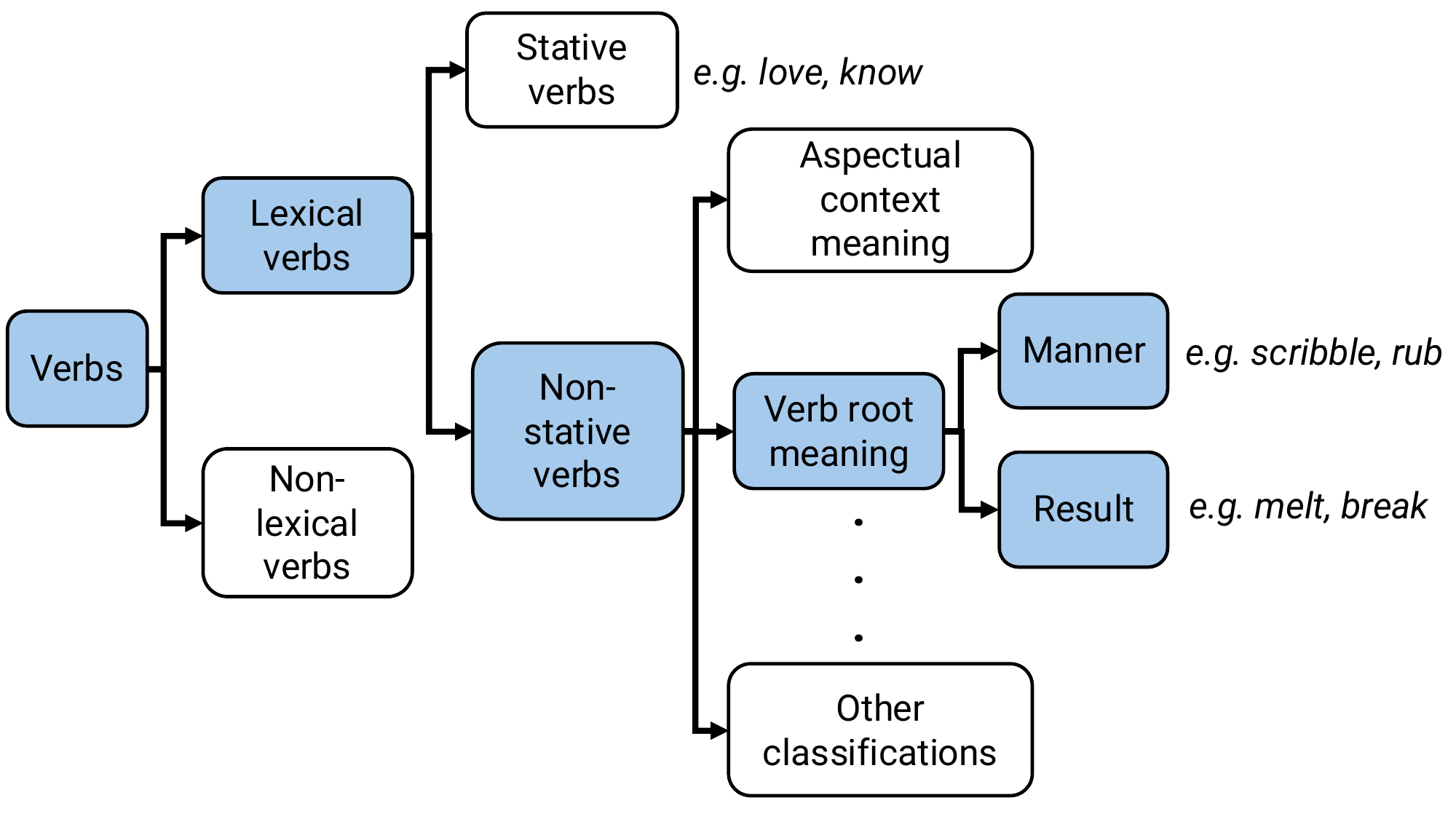}
	\caption{Hierarchy of verb classification, with manner and result verbs as subdivisions of non-stative verbs}
	\label{fig:verb_hierarchy}
\end{figure}

\noindent Figure \ref{fig:verb_hierarchy} shows the hierarchy of verb classifications relevant to our proposed task. At a high level, lexical verbs can be categorized into \textbf{stative} and \textbf{non-stative} verbs.  
\begin{itemize}[leftmargin=*]\setlength\itemsep{0em}
    \item \textbf{Stative verbs} describe a continuous or unchanging state rather than an action or event, e.g.~\emph{love} in the sentence “She loves her dog,” 
    \item \textbf{Non-stative verbs}, on the other hand, describe actions or events that unfold over time and can lead to changes in state.
\end{itemize}

Non-stative verbs can be further classified based on different linguistic properties, such as \textbf{aspectual features} (e.g., telicity, durativity) and \textbf{argument realization patterns} (e.g., causative-inchoative alternation),etc. However, a fundamental classification based on the \textbf{inherent meaning stored in the verb root} is the difference between manner verbs and result verbs \cite{levin1991wiping, hovav2010reflections, levin2008constraint}; This distinction plays a significant role in both language acquisition \cite{gentner2001individuation} and the way verbs encode event semantics.
\begin{itemize}[leftmargin=*]\setlength\itemsep{0em}
    \item \textbf{Manner verbs} specify \textit{how} an action is performed but do not encode its outcome (e.g., \textit{scribble, rub, sweep, flutter}).
    \item \textbf{Result verbs} specify \textit{what} change or outcome occurs, without specifying how the action is carried out (e.g., \textit{clean, melt, fill, arrive}).
\end{itemize}
Unlike classifications such as telicity, which are determined at the clause level \cite{friedrich2017classification}, the manner/result distinction is typically analyzed as a property of the verb root \cite{levin2008constraint}, meaning that it is expected to remain relatively stable across contexts.

\subsection{Illustrating the difference between manner and result verbs}
To understand this complementarity, consider the following pair of sentences:
\begin{enumerate}\setlength\itemsep{0mm}
    \item \emph{Anna shoveled the snow.} 
    \item \emph{Anna cleared the snow.}
\end{enumerate}

In (1), the verb \emph{shoveled} focuses on \textit{how} the action was performed, i.e. the process of moving the snow with a shovel, but does not guarantee that the snow was removed.
In contrast, in (2), the verb \emph{cleared} encodes the outcome, that the snow was removed, but does not specify how Anna accomplished this (she could have used a shovel, a snowblower, or even melted it).
This distinction is crucial because it shows that result verbs inherently encode a outcome, while manner verbs focus on the process. 
One way to test whether a verb encodes a result or manner is by using the \textit{denying the result} diagnostic test \cite{hovav2010reflections}. If the sentence remains logical, the verb does not inherently encode a result:
\begin{quote}
    \emph{Anna shoveled the snow, but the snow is still there.} (logical)
\end{quote}
Since this sentence makes sense, we can infer that ``\emph{shovel}'' does not encode a result; it only describes the action. Thus even though real-world knowledge might suggest that performing an action in a certain way will typically lead to a result, this is not always true. The \textbf{core meaning of a verb remains stable across different contexts}. 
However, trying the same test with a result verb leads to contradiction:
\begin{quote}
    \emph{Anna cleared the snow, but the snow is still there.} (contradiction)
\end{quote}

\section{Manner and Result Verb Diagnostics}\label{sec:diagnostics}
To effectively transfer the knowledge of result and manner heuristics into an LLM annotator, it is essential to identify the linguistic features that reliably distinguish them. Since the manner/result distinction is inherent to the verb root rather than being compositionally determined, much of this semantic information is encoded within the verb itself. However, sentence structure also offers useful cues, as manner and result verbs occur in complementary syntactic environments.
In particular:
\begin{itemize}[leftmargin=*]\setlength\itemsep{0em}
    \item Manner verbs frequently occur without a direct object.
    \item Result verbs typically require an object to specify the entity undergoing change.
    \item Only result verbs consistently participate in causative/inchoative alternations.
\end{itemize}

Below, we present these sentence formation diagnostics that linguistic researchers have leveraged for result and manner verb identification.

\subsection{Sentence formation diagnostics}
\paragraph{Diagnostic 1: Object omission}\label{sec:diagnostic1}
Manner verbs can appear without a direct object, whereas result verbs typically require one \cite{hovav2010reflections}. Consider the following examples:

\begin{itemize}[leftmargin=*]\setlength\itemsep{0em}
    \item Manner verb: \emph{Anna wept all day.} (Acceptable without an object)
    \item Result verb: \emph{The child broke \_ ?} (Unacceptable without an object)
\end{itemize}

This suggests that manner verbs describe an action that can occur independently, whereas result verbs typically requiring an affected entity.

\paragraph{Diagnostic 2: causative/inchoative alternation}\label{sec:diagnostic2}
The causative/inchoative alternation refers to a pattern in which a verb appears both in a causative form (with an explicit agent) and an inchoative form (where the event occurs spontaneously without an agent)\cite{hovav2010reflections, beavers2012manner, levin1991wiping}. This alternation serves as a reliable test for result verbs, as manner verbs rarely allow such transformations.

\begin{itemize}[leftmargin=*]\setlength\itemsep{0em}
    \item Result Verb:
    \begin{itemize}
        \item \emph{Causative:} \emph{The child broke the vase.} (An agent explicitly causes the event.)
        \item \emph{Inchoative:} \emph{The vase broke.} (The event occurs without an explicit agent.)
    \end{itemize}
    \item Manner Verb:
    \begin{itemize}
        \item \emph{Causative (transitive):} \emph{John wiped the table.}
        \item \emph{Inchoative (intransitive):} \emph{The table wiped.} (Ungrammatical)
    \end{itemize}
    
    Unlike result verbs, manner verbs describe a process but do not inherently encode an endpoint. As a result, they resist appearing in inchoative constructions.
\end{itemize}

\subsection{Semantic Diagnostics: beyond syntactic patterns}
While the above syntactic tests provide useful heuristics, they are not always sufficient for classification. Certain verbs such as \emph{climb}, and \emph{cut} resist strict categorization due to polysemy or context-dependent interpretations \cite{levin2008constraint, beavers2012manner}. To address this, researchers have therefore investigated \textbf{semantic properties} that further refine the manner/result distinction.

\paragraph{Diagnostic 3: Telicity}\label{sec:diagnostic3}
Telicity refers to whether a verb's action has a natural endpoint or goal. A verb is \emph{telic} if it describes an action that reaches completion, such as \emph{build} or \emph{paint} (\emph{She built a house.}, \emph{He painted a portrait.}). These actions have a defined conclusion. In contrast, a verb is \emph{atelic} when the action is ongoing, lacks a specific endpoint, or its completion is uncertain, as seen with verb like \emph{sleep} (\emph{They slept peacefully}). \citet{dowty2012word, levin1991wiping, krifka1992thematic} observed a correlation between result verbs and telicity. However, while result verbs involving two-point changes (e.g., arrive, reach, die, crack, find) are necessarily telic, result verbs describing degree achievements verbs (cooled, heat) are not strictly telic. Consider the shift in telicity with a time modifier.
\begin{itemize}[leftmargin=*]\setlength\itemsep{0em}
    \item \emph{The dryer dried the clothes for two hours} \\(Atelic: no clear endpoint)
    \item \emph{The dryer dried the clothes in two hours} \\(Telic: the drying is completed)
\end{itemize}

\paragraph{Diagnostic 4: scalar vs. non-scalar changes}\label{sec:diagnostic4}
\citet{hovav2010reflections} proposed that the distinction between \textbf{scalar} and \textbf{non-scalar} changes provides a strong basis for differentiating manner and result verbs. Since both verb types denote dynamic events, they inherently involve a change \cite{dowty2012word}; however, the nature of that change differs. Result verbs are characterized by changes that occur along a measurable scale, either as a two-point change (e.g., break) or as a gradable change (e.g., melt). In contrast, manner verbs involve non-scalar changes that cannot be readily quantified along a single dimension. For example, the action described by the verb \emph{flap} entails a complex, multidimensional movement that is not easily measurable.\\
Result verbs thus describe changes along a measurable scale, meaning the event involves a progression toward a defined endpoint.
    \begin{itemize}[leftmargin=*]\setlength\itemsep{0em}
        \item Two-point scale (binary change): \\ \emph{break, die, arrive}
        \item Gradable scale (continuous change):\\ \emph{melt, cool, widen}
    \end{itemize}

Manner verbs describe non-scalar changes, where the event unfolds without a well-defined trajectory.
    \begin{itemize}[leftmargin=*]\setlength\itemsep{0em}
        \item Example: \emph{flap, jog, scribble}-these actions involve repeated or multidimensional motion rather than progression toward an endpoint.
    \end{itemize}

This distinction supports \citet{levin2008constraint} hypothesis of manner-result complementarity, which posits that a single action cannot simultaneously encode both a scalar and non-scalar change. 

\subsection{Implications for LLM annotation}
The manner/result distinction is semantically encoded (as part of the verb meaning), but syntactic diagnostics contribute in testing participation of a verb in a particular category using sentence structure. These distinctions are integrated into our approach by structuring our prompt designs around the sentence formation rules (diagnostic tests 1 and 2) and semantic features (diagnostic tests 3 and 4).

\section{Methodology}\label{sec:method}
In this section, we describe the end-to-end pipeline, including (i) annotation of training data for manner/result labels, (ii) training of the tagging model, and (iii) construction of held-out (gold-standard) evaluation datasets.

\begin{figure*}[h]
	\centering
	\includegraphics[width=0.85\linewidth]{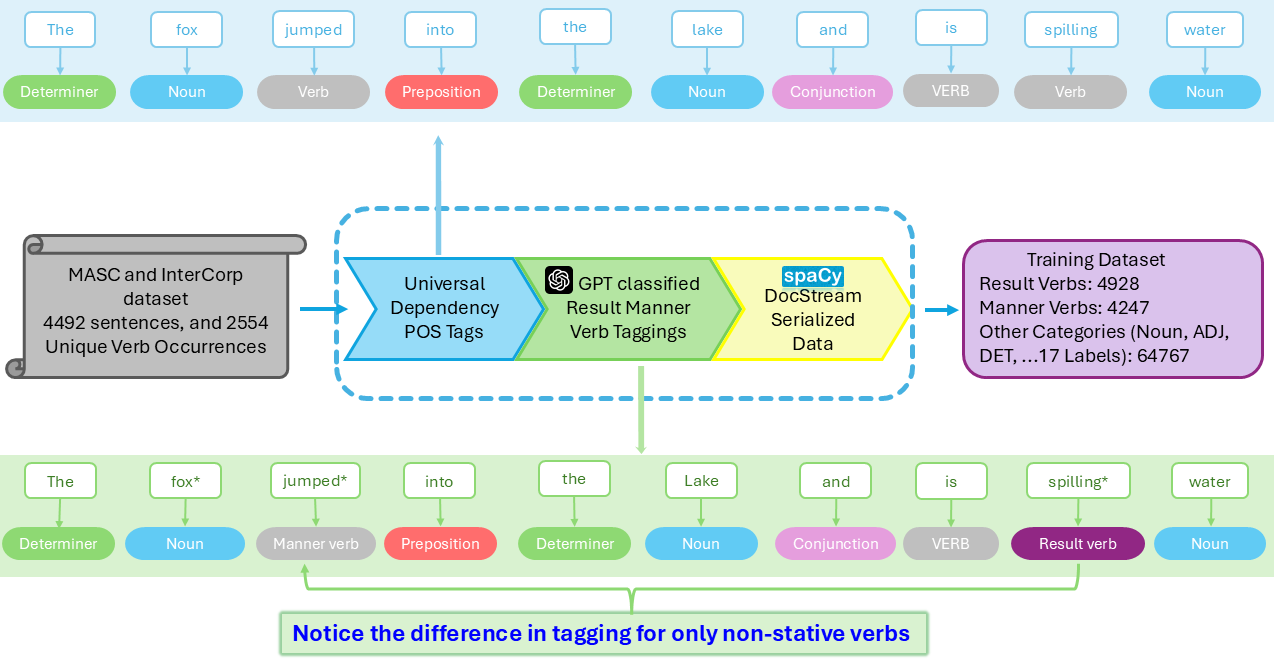}
	\caption{Overview of our data generation pipeline.}
	\label{fig:LLMpipeline}
\end{figure*}

As explained in the Contributions list from Section \ref{sec:intro}, to the best of our knowledge, this is the first attempt to computationally annotate and classify texts using the manner/result constructs. For this reason, there are no known annotated datasets useful for training a computational model. Hence, to address this challenge, we resorted to LLMs, to assist in creating a large, annotated dataset with result and manner verb labels. \citet{he2023annollm,zhang2023llmaaa} showed that with structured prompts and few-shot examples, LLMs can effectively mimic human annotations for various NLP tasks.


\subsection{LLM-Based training data annotation}

 For this task, we compile the sentences from MASC and InterCorp dataset consisting of 4,492 sentences and 2,554 unique verb occurrences. Next, using our expert-guided prompts, we use the GPT-4o model to identify the non-stative verbs in each sentence and classify them based on our manner-result diagnostic framework. The rules for designing the two separate prompts for GPT-4o, where each focuses on a different aspect of verb classification, are described:
 
 \textbf{Prompt 1 (semantic properties):} checks for scalar vs. non-scalar change information embedded within verb root. The two major rules driving Prompt 1 are shown in Figures \ref{fig:resultmanner-def} and \ref{fig:verb-root-rule}.
 
\textbf{Prompt 2 (sentence structure):} emphasizes possible sentence formation patterns, including object omission and causative/inchoative alternations. Due to space constraints, the governing rules for Prompt 2 is presented in the Appendix \ref{sec:appendixB}. \\

\begin{figure}[ht!p]
\centering
\begin{mdframed}
\scriptsize
\noindent
\\
\textbf{Manner verbs (manner):}\\
\textit{Definition}: These verbs encode the \textit{how} of an action, focusing on the method or pattern without specifying an outcome.\\
\textit{Semantic Basis}: These verb often involve \textbf{nonscalar} or \textbf{complex} actions that are often multidimensional (e.g., the specific pattern of leg movements while jogging which is complex and a culmination of multiple actions.)\\
\textit{Usage}: ``She \textbf{ran} towards the market.'' (Focus on how she went to the market)
\\[1em]
\textbf{Result verbs (result):}\\
\textit{Definition}: These verbs encode the \textit{outcome} or resultant state that follows from an action.\\
\textit{Semantic Basis}: Involve \textbf{scalar} changes that occur along a defined scale (e.g., temperature increasing, distance decreasing).\\
\textit{Usage}: ``He \textbf{melted} the ice.'' (Focus on the fact that \textbf{melting} alone stores all the information that something has changed form)
\end{mdframed}
\caption{Result Manner Verb Definition}
\label{fig:resultmanner-def}
\end{figure}

\begin{figure}[ht!p]
\centering
\begin{mdframed}
\scriptsize
\noindent
\textbf{Verb Root Classification}\\
\textbf{Definition:} The verb has to be classified based on its primary lexical meaning and the inherent information that the verb independently encodes irrespective of the context.\\[0.5em]
\textbf{Example: \emph{``Wipe''}}\\
\textit{Sentence:} ``He \textbf{wiped} the table \textbf{clean}.'' \\
\quad The verb \emph{wipe} primarily indicates the manner of cleaning; the resulting state (\textit{``clean''}) is introduced by the adjective \emph{clean}, not by \emph{wipe} itself. Therefore, the verb \emph{wipe} remains a Manner Verb because the outcome (i.e., making something clean) is not inherently encoded by the verb’s own meaning.\\[0.5em]
\textbf{Past Information in Verb Classification:}\\
Manner Verbs inherently encode the way an action was performed in past instances, while Result Verbs do not.\\[0.5em]
\textit{Example 1: \textbf{Cook} (Result Verb)}\\
\quad Sentence: ``She \textbf{cooked} chicken for him.''\\
\quad Rewriting it as: ``She \textbf{cooked} chicken for him again.''
\quad The word \textbf{cook} does not provide information about how the food was cooked before, it can be grilled, sautéed, etc. \\[0.5em]
\textit{Example 2: \textbf{Sauté} (Manner Verb)}\\
\quad Sentence: ``She \textbf{sautéed} chicken for him.''\\
\quad Rewriting it as: ``She \textbf{sautéed} chicken for him again.''
\quad The verb \textbf{sauté} stores the information that the chicken was previously also cooked using sautéeing.
\end{mdframed}
\caption{Verb Root Classification}
\label{fig:verb-root-rule}
\end{figure}

The two prompts were presented to the Gpt-4o LLM.
Prompt 1 yielded 4,928 result verbs, 4,247 manner verbs, and 64,767 words tagged with other categories such as nouns, determiners, pronouns, etc. Prompt 2 yielded 4,813 result verbs, 4,354 manner verbs, and 64,775 words tagged into other categories.


\subsection{Approaching the problem as part-of-speech (POS) tagging}
Since our task involves both verb classification and detection them in a sentence, we adopt a sequence-tagging approach, similar to part-of-speech (POS) tagging, rather than formulating it as a binary classification task. This enables us to identify non-stative verbs, since modal and auxiliary verbs are readily identifiable using syntactic structures.

The advantages of taking a sequence-tagging approach include:
\begin{enumerate}
    \item Explicit identification of non-stative verbs: By tagging all the words in a sentence, we can reduce the final error by isolating and classifying only the non-stative verbs, thus avoiding any misclassification of auxiliary and modal verbs (e.g., \emph{can, might, have, be}).
    \item Facilitates our ultimate goal in child language research applications: Our model can be directly integrated into the child language research pipeline where most often the goal is to scan through the complete sentences spoken by a child, and identify the number of result and manner verbs. Tagging only the non-stative verbs eliminates an additional step to filter any stative and non-lexical verbs.
\end{enumerate}

Figure \ref{fig:LLMpipeline} illustrates the sequence-tagging based data generation pipeline.  First, we tag each sentence using any standard POS tagger. For example, the sentence "\emph{The fox jumps into the lake and is spilling water}" is initially tagged as:\\
"\emph{The} (DT) \emph{fox} (NN) \emph{jumps} (VB) \emph{into} (IN) \emph{the} (DT) \emph{lake} (NN) \emph{and} (CC) \emph{is} (VB) \emph{spilling} (VB) \emph{water} (NN)."\\
Next, we update the tagging for non-stative verbs using GPT-4o \cite{achiam2023gpt}, classifying them as either result or manner verbs. The modified tagged dictionary:\\
"\emph{The} (DT) \emph{fox} (NN) \emph{jumps} (manner) \emph{into} (IN) \emph{the} (DT) \emph{lake} (NN) \emph{and} (CC) \emph{is} (VB) \emph{spilling} (result) \emph{water} (NN)."\\
This process is applied to all sentences, and finally compiled to create the training set.

\subsection{Curation of gold-standard test data}\label{sec:testDataCuration}
We evaluate our models on three held-out gold-standard datasets. The first, the \emph{Linguists verb-root} dataset, contains 83 verbs (34 result, 49 manner) compiled from prior work on lexical semantics and verb-root classification \citep{levin2008constraint, hovav2010reflections, beavers2012manner, levin1991wiping}. The second, the \emph{Psycholinguistic verb-root} dataset, comes from \citet{horvath2022difference}, who annotated 77 MacArthur-Bates CDI verbs (36 result, 41 manner).

Because these datasets together covered only 151 of 487 VerbNet classes, we created a third set with the help of an expert linguist. Guided by VerbNet, we constructed 200 new sentences spanning 346 classes; the expert labeled 48 as result, 62 as manner, 23 as stative, and 67 as unsure. We refer to this as the \emph{Expert-annotated verb-root} dataset (see Appendix~\ref{sec:appendixA}).

All 3 datasets are distinct from the MASC training data and are used only for held-out evaluation.


\section{Computational Modeling}
\label{sec:compmodel}

This section outlines our computational approach for classifying verbs according to both \textit{manner/result} and \textit{stative/non-stative} properties. 
\begin{figure*}[h]
	\centering
	\includegraphics[width=0.9\linewidth]{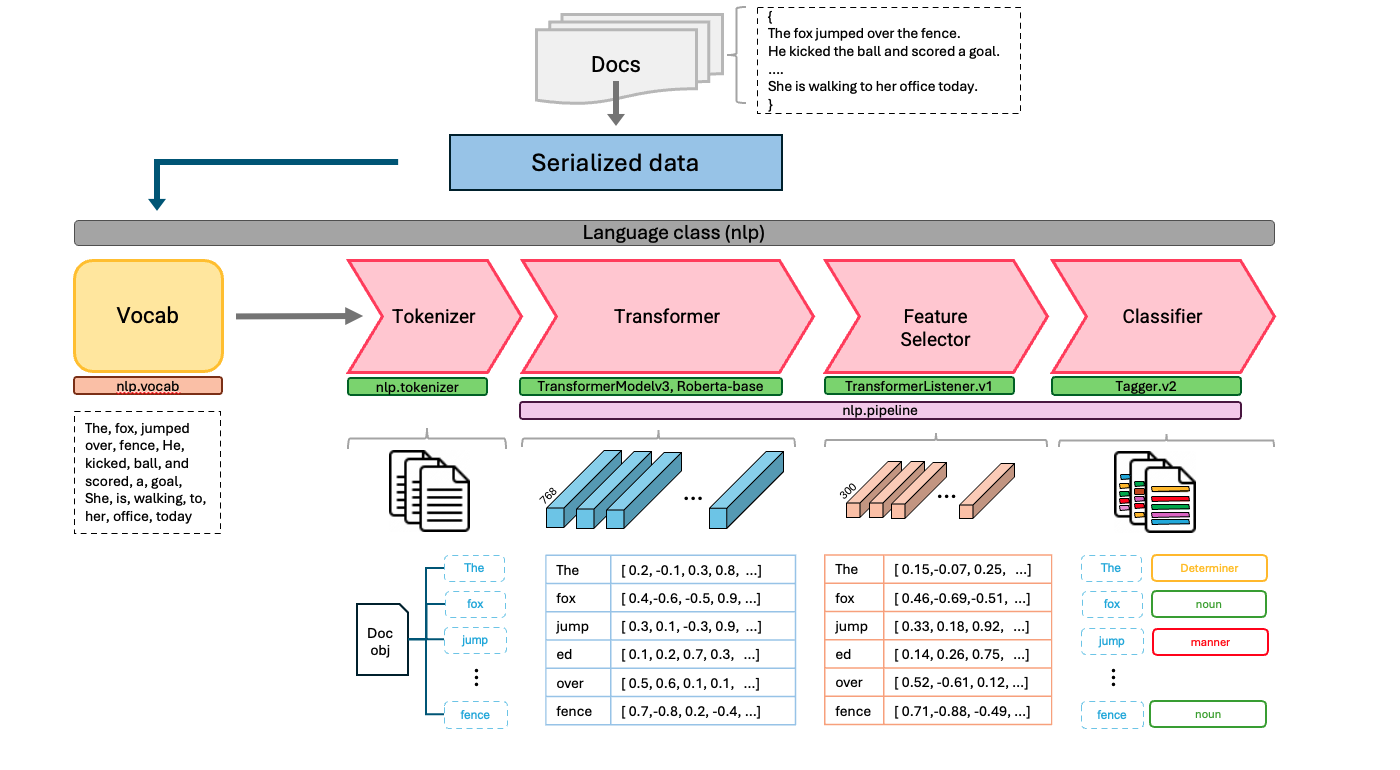}
	\caption{Overview of model architecture}
	\label{fig:arch}
\end{figure*}

\subsection{Model architecture}
\label{ssec:modelarch}

Our tagging pipeline is implemented using a spaCy wrapper \cite{spacy2} and follows a sequence of components as shown in \autoref{fig:arch}: (1) a tokenizer, (2) fine tuning a pre-trained transformer-based feature extractor, (3) a feature selector (pooling layer), and (4) a classification head.

\paragraph{Tokenizer.}
Byte Pair Encoding Tokenization \cite{sennrich2015neural} strategy segments raw text into tokens, and matches with our downstream RoBERTa model default tokenization strategy.
\paragraph{Transformer.}
We employ \textbf{RoBERTa-base} model (125 million parameters) as the backbone of our pipeline, which encodes each token - in conjunction with its context - into a contextualized representation. 
\paragraph{Feature Selector.}
To reduce subword embeddings to a single vector per token, we apply mean pooling (\texttt{reduce\_mean.v1}). 
\paragraph{Classifier.}
We use label smoothing ($0.05$) to predict token-level labels for default parts-of-speech tagging (17) plus two new labels (result and manner) Each token’s pooled embedding is projected into logits corresponding to these classes, and the model is optimized via cross-entropy loss.

\subsection{Feature representation}
\label{ssec:features}

\paragraph{Contextual embeddings.}
Tokens are generated using BPE tokenizer that sequences via pretrained \textit{RoBERTa-base} vocabulary. This aids in capturing syntactic signals.

\paragraph{Token-Level pooling.}
Mean pooling operation over subword embeddings yields 768-dimensional vectors  representing token-level features. A feature selector (\texttt{TransformerListener}) is applied to remove redundant information, reducing them to 300-dimensional representations, retaining semantic and syntatic features.

\subsection{Training procedure}
\label{ssec:training}

\paragraph{Hyperparameters.}
We train the model using Adam with learning rate $= 5 \times 10^{-5}$, $\beta_1=0.9, \beta_2=0.999$, weight decay (L2) $= 0.01$, gradient clipping $= 1.0$ and batch size $= 128$. 

The model is trained for up to 20,000 steps, with evaluation every 200 steps. A patience of 1,600 steps is used to halt training if the validation accuracy fails to improve. This setup balances thorough exploration of the parameter space with computational efficiency.

All experiments run using a word-based batcher and compounding batch sizes (start=100, stop=1000, compound=1.001) on a single GPU (NVIDIA RTX A6000) for 25 minutes training time.  The final checkpoint is selected based on the highest tagging accuracy on our gold annotated dataset.

\setlength\extrarowheight{1mm}
\begin{table*}[h!tp]
\centering
\begin{tabular}{lccccccc}
\hline
 & Acc. & F$_1$ & Precision & Recall & F$_1$ & Precision & Recall\\
 & & (result) & (result) & (result) & (manner) & (manner) & (manner)\\
\hline
\multicolumn{8}{c}{\textbf{Model 1 (Trained using Prompt 1)}} \\ \hline
Linguistic dataset & 0.94 & 0.93 & 0.89 & 0.97 & 0.95 & 0.98 & 0.92 \\ 
Psycholinguistic dataset & 0.90 & 0.88 & 1.00 & 0.78 & 0.91 & 0.84 & 1.00 \\ 
Expert-annotated dataset & 0.86 & 0.85 & 0.84 & 0.85 & 0.88 & 0.89 & 0.87 \\ 
\hline
\multicolumn{8}{c}{\textbf{Model 2 (Trained using Prompt 2)}} \\ \hline
Linguistic dataset & 0.94 & 0.93 & 0.91 & 0.94 & 0.95 & 0.96 & 0.94 \\ 
Psycholinguistic dataset & 0.84 & 0.80 & 1.00 & 0.67 & 0.87 & 0.77 & 1.00 \\ 
Expert-annotated dataset & 0.81 & 0.80 & 0.82 & 0.77 & 0.84 & 0.84 & 0.84\\ 
\hline
\end{tabular}
\caption{Comparison of Model 1 and Model 2 on different datasets.}
\label{tab:model_comparison}
\end{table*}

\newpage
\section{Experiments and Results}\label{sec:expts}
We evaluate our models on the three gold-standard datasets described in Section \ref{sec:testDataCuration}, the Linguist, Psycholinguists and Expert-annotated verb root datasets.

\paragraph{Quantitative Results}
We trained our model using annotations generated from two distinct prompts -one emphasizing the semantic properties of verbs and the other focusing on sentence structure. 
Table~\ref{tab:model_comparison} presents model performance across multiple datasets, highlighting accuracy, F1-score, precision, and recall for result and manner verbs.
\begin{itemize}
    \item Model 1 consistently outperforms Model 2 achieving equal or higher accuracy across all three datasets. 
    \item The Linguistics dataset performed the best among all three test datasets and across the two prompts. This is likely due to the fact that we constructed our governing prompt rules based on information gleaned from the papers from which that dataset was culled.
    \item Model 1 shows weaker recall (0.67) for result verbs on the Psycholinguistic dataset, indicating higher misclassification rates. Inspection of the disagreements suggests that some verbs in this dataset (e.g., \textit{paint}, \textit{dump}, \textit{drink}) may be borderline cases, for which psycholinguistic annotations and our linguistically guided framework assign different labels. Since the dataset in \citet{horvath2022difference} was annotated for developmental research purposes, these mismatches may reflect differences in annotation criteria across domains rather than simply model failure.\citet{horvath2019syntactic} indicated in their paper that the authors annotated the verbs themselves. 
\end{itemize}
The fact that Model 1 performs better than Model 2 suggests that understanding the semantic information inherent in verb roots is more crucial than analyzing sentence structure, for this verb categorization task.

\section{Developmental Use Case}
We illustrate the utility of our approach through its application to a longitudinal developmental dataset of parent-child interactions involving typically developing (TD) toddlers and Late Talkers (LTs). In this use case, transcripts from the CHILDES Clinical English Ellis Weismer corpus \citep{weismer2013fast} are processed with our classifier to identify manner and result verbs in caregiver and child speech at 30 months, yielding speaker-level measures such as verb types, tokens, and manner-to-result ratios. These measures can then be related to later language outcomes at 42 and 66 months, including MLU, TTR, and IPSyn scores. IPSyn was measured following \citet{scarborough1990index}, TTR was included as a standard index of lexical diversity \citep{hess1986sample}, and mean length of utterance (MLU) was derived from examiner-child language samples at two time points to index later grammatical development \citep{weismer2013fast}.

This use case is motivated by prior developmental findings suggesting that manner and result verbs may differ in their relation to language growth. For example, \citet{horvath2022difference} report that TD children’s vocabularies contain relatively more manner verbs, whereas Late Talkers’ vocabularies contain relatively more result verbs; children who produce more manner verbs also tend to produce more verbs overall. At the same time, broad measures of parental input have not consistently distinguished the language environments of TD children and Late Talkers, suggesting that finer-grained semantic properties of the input may also be informative \citep{d2006prosodic,naigles1998some}.

We view this as an example of the kind of developmental analysis that scalable manner/result classification supports. Rather than relying only on coarse measures of input quantity, researchers can use the present tool to examine whether the semantic composition of caregiver and child verb use is related to later language outcomes. In this sense, the classifier functions as a corpus-based measurement tool that may help support richer analyses of early verb learning and developmental variation.

This use case is intended as an illustration of research utility rather than a clinical application. Although the tool supports corpus-based measurement of a theoretically motivated semantic distinction, further validation will be needed before drawing stronger conclusions about diagnostic use or generalization across developmental datasets \citep{verhage2020collaboration,conti2018education}.


\section{Conclusion}\label{sec:concl}
We present a computational approach to identifying manner and result verbs in context. By using LLM-generated annotations, we expand coverage from 151 to 436 VerbNet classes and train a RoBERTa-based classifier on this distinction.

The model achieves up to 89.6\% average accuracy across three gold-standard evaluation sets (with annotations by expert linguists). Our results suggest that semantic properties of non-stative verb roots contribute more to this task than sentence structure alone, supporting the value of linguistically informed modeling for event-structure classification.

Future work will test the approach on more diverse data, extend it to other languages, and further explore its use in developmental language research. At present, we see the model as a tool for corpus-based analysis which can have developmental relevance, rather than as a direct clinical or diagnostic decision-making tool.



\newpage
\section*{Limitations}
The following section illustrates some of the current limitations of the proposed research:
\begin{itemize}
    \item In this work, although we have identified comprehensive sets of manner/results verb diagnostics, and have used these to construct intelligent prompt for generating our training data, \emph{we did not consider polysemous verbs and subtle alternations of verbs}.

\item While LLMs perform well in verb categorization, they rely on statistical associations rather than linguistic principles, and this could lead to inconsistencies. \emph{When a random sampling of the resulting annotated data was ``spot-checked'' by an expert, the LLM annotations were \underline{not} 100\% accurate}.  

\item Subsequent analyses by \citet{beavers2012manner} noted that certain verbs exhibit both manner and result properties. For instance, the verb \emph{guillotine}, and \emph{drowned} explicitly convey the manner of killing (i.e., how the action is performed) while also implying the resultant state (i.e., that the person is killed). Similar behavior is observed with certain cooking verbs such as \emph{braise}, \emph{sauté}, and \emph{poach}. However, \emph{for our analysis in this work, we focused only on the manner \underline{or} result aspect of non-stative verbs}.

\item A critical challenge in this work was the scarcity of expertise in the research area, with only a handful of specialists available. We therefore relied mainly on one expert to create our gold-standard expert annotation and \emph{we were unable to obtain inter-rater reliability}. 
\end{itemize}

\section*{Ethical Considerations}
This work raises several ethical considerations relevant to computational tools for semantic annotation.

\begin{itemize}
    \item Large language models may reproduce linguistic and cultural biases present in their training data. Consequently, our annotation pipeline may be less accurate for speakers whose language use is underrepresented in standard corpora, including speakers of regional or non-standard varieties of English.

    \item We view the present system as a research tool rather than a clinical or diagnostic instrument. Any future use in developmental or clinical contexts would require substantial additional validation across populations, settings, and language varieties.

    \item Since the current data are drawn primarily from standard English sources, the model may not generalize equally well to all communities. Expanding evaluation to more diverse dialects and speech contexts is therefore an important direction for future work.
\end{itemize}

\section*{Broader Impacts}
This work has potential broader impacts across developmental language research, linguistics, and computational modeling.

\begin{itemize}
    \item For developmental research, scalable measurement of manner and result verbs may support more fine-grained analyses of early language development, including studies of late talkers and children at risk for persistent language difficulties. Because later outcomes among children with early language delay are heterogeneous, tools that make theoretically motivated semantic distinctions measurable in larger corpora may help researchers better characterize variation in children’s early vocabularies and language input \citep{catts2012prevalence,conti2018education}.

    \item For computational linguistics, this work provides an example of how linguistic theory and domain expertise can be incorporated into annotation pipelines and downstream modeling. In particular, the use of linguistically informed prompts illustrates one possible strategy for generating training data in tasks where large gold-standard semantic resources are not yet available.
\end{itemize}

\newpage
\bibliography{custom}

\begin{thebibliography}{33}
\providecommand{\natexlab}[1]{#1}

\bibitem[{Achiam et~al.(2023)Achiam, Adler, Agarwal, Ahmad, Akkaya, Aleman,
  Almeida, Altenschmidt, Altman, Anadkat et~al.}]{achiam2023gpt}
Josh Achiam, Steven Adler, Sandhini Agarwal, Lama Ahmad, Ilge Akkaya,
  Florencia~Leoni Aleman, Diogo Almeida, Janko Altenschmidt, Sam Altman,
  Shyamal Anadkat, et~al. 2023.
\newblock Gpt-4 technical report.
\newblock \emph{arXiv preprint arXiv:2303.08774}.

\bibitem[{Beavers and Koontz-Garboden(2012)}]{beavers2012manner}
John Beavers and Andrew Koontz-Garboden. 2012.
\newblock Manner and result in the roots of verbal meaning.
\newblock \emph{Linguistic inquiry}, 43(3):331--369.

\bibitem[{Brown et~al.(2019)Brown, Bonn, Gung, Zaenen, Pustejovsky, and
  Palmer}]{brown2019verbnet}
Susan~Windisch Brown, Julia Bonn, James Gung, Annie Zaenen, James Pustejovsky,
  and Martha Palmer. 2019.
\newblock Verbnet representations: Subevent semantics for transfer verbs.
\newblock In \emph{Proceedings of the First International Workshop on Designing
  Meaning Representations}, pages 154--163.

\bibitem[{Catts et~al.(2012)Catts, Compton, Tomblin, and
  Bridges}]{catts2012prevalence}
Hugh~W Catts, Donald Compton, J~Bruce Tomblin, and Mindy~Sittner Bridges. 2012.
\newblock Prevalence and nature of late-emerging poor readers.
\newblock \emph{Journal of educational psychology}, 104(1):166.

\bibitem[{Conti-Ramsden et~al.(2018)Conti-Ramsden, Durkin, Toseeb, Botting, and
  Pickles}]{conti2018education}
Gina Conti-Ramsden, Kevin Durkin, Umar Toseeb, Nicola Botting, and Andrew
  Pickles. 2018.
\newblock Education and employment outcomes of young adults with a history of
  developmental language disorder.
\newblock \emph{International journal of language \& communication disorders},
  53(2):237--255.

\bibitem[{DeRose(1988)}]{derose1988grammatical}
Steven~J. DeRose. 1988.
\newblock \href {https://aclanthology.org/J88-1003/} {Grammatical category
  disambiguation by statistical optimization}.
\newblock \emph{Computational Linguistics}, 14(1):31--39.

\bibitem[{D'Odorico and Jacob(2006)}]{d2006prosodic}
Laura D'Odorico and Valentina Jacob. 2006.
\newblock Prosodic and lexical aspects of maternal linguistic input to
  late-talking toddlers.
\newblock \emph{International Journal of Language \& Communication Disorders},
  41(3):293--311.

\bibitem[{Dowty(2012)}]{dowty2012word}
David~R Dowty. 2012.
\newblock \emph{Word meaning and Montague grammar: The semantics of verbs and
  times in generative semantics and in Montague's PTQ}, volume~7.
\newblock Springer Science \& Business Media.

\bibitem[{ek~{\v{C}}erm{\'a}k and Rosen(2012)}]{ek2012case}
Franti ek~{\v{C}}erm{\'a}k and Alexandr Rosen. 2012.
\newblock The case of intercorp, a multilingual parallel corpus.
\newblock \emph{International Journal of Corpus Linguistics}, 17(3):411--427.

\bibitem[{Friedrich and Gateva(2017)}]{friedrich2017classification}
Annemarie Friedrich and Damyana Gateva. 2017.
\newblock Classification of telicity using cross-linguistic annotation
  projection.
\newblock In \emph{Proceedings of the 2017 Conference on Empirical Methods in
  Natural Language Processing}, pages 2559--2565.

\bibitem[{Friedrich et~al.(2016)Friedrich, Palmer, and
  Pinkal}]{friedrich2016situation}
Annemarie Friedrich, Alexis Palmer, and Manfred Pinkal. 2016.
\newblock Situation entity types: automatic classification of clause-level
  aspect.
\newblock In \emph{Proceedings of the 54th Annual Meeting of the Association
  for Computational Linguistics (Volume 1: Long Papers)}, pages 1757--1768.

\bibitem[{Friedrich et~al.(2022)Friedrich, Xue, and Palmer}]{friedrich2022kind}
Annemarie Friedrich, Nianwen Xue, and Alexis Palmer. 2022.
\newblock A kind introduction to lexical and grammatical aspect, with a survey
  of computational approaches.
\newblock \emph{arXiv preprint arXiv:2208.09012}.

\bibitem[{Gentner and Boroditsky(2001)}]{gentner2001individuation}
Dedre Gentner and Lera Boroditsky. 2001.
\newblock Individuation, relativity, and early word learning.
\newblock \emph{Language acquisition and conceptual development}, 3:215--256.

\bibitem[{Hadley et~al.(2016)Hadley, Rispoli, and Hsu}]{hadley2016toddlers}
Pamela~A Hadley, Matthew Rispoli, and Ning Hsu. 2016.
\newblock Toddlers' verb lexicon diversity and grammatical outcomes.
\newblock \emph{Language, speech, and hearing services in schools},
  47(1):44--58.

\bibitem[{He et~al.(2023)He, Lin, Gong, Jin, Zhang, Lin, Jiao, Yiu, Duan, Chen
  et~al.}]{he2023annollm}
Xingwei He, Zhenghao Lin, Yeyun Gong, Alex Jin, Hang Zhang, Chen Lin, Jian
  Jiao, Siu~Ming Yiu, Nan Duan, Weizhu Chen, et~al. 2023.
\newblock Annollm: Making large language models to be better crowdsourced
  annotators.
\newblock \emph{arXiv preprint arXiv:2303.16854}.

\bibitem[{Hess et~al.(1986)Hess, Sefton, and Landry}]{hess1986sample}
Carla~W Hess, Karen~M Sefton, and Richard~G Landry. 1986.
\newblock Sample size and type-token ratios for oral language of preschool
  children.
\newblock \emph{Journal of Speech, Language, and Hearing Research},
  29(1):129--134.

\bibitem[{Honnibal and Montani(2017)}]{spacy2}
Matthew Honnibal and Ines Montani. 2017.
\newblock {spaCy 2}: Natural language understanding with {B}loom embeddings,
  convolutional neural networks and incremental parsing.
\newblock To appear.

\bibitem[{Horvath et~al.(2022)Horvath, Kueser, Kelly, and
  Borovsky}]{horvath2022difference}
Sabrina Horvath, Justin~B Kueser, Jaelyn Kelly, and Arielle Borovsky. 2022.
\newblock Difference or delay? syntax, semantics, and verb vocabulary
  development in typically developing and late-talking toddlers.
\newblock \emph{Language Learning and Development}, 18(3):352--376.

\bibitem[{Horvath et~al.(2019)Horvath, Rescorla, and
  Arunachalam}]{horvath2019syntactic}
Sabrina Horvath, Leslie Rescorla, and Sudha Arunachalam. 2019.
\newblock The syntactic and semantic features of two-year-olds’ verb
  vocabularies: A comparison of typically developing children and late talkers.
\newblock \emph{Journal of Child Language}, 46(3):409--432.

\bibitem[{Hovav and Levin(2010)}]{hovav2010reflections}
Malka~Rappaport Hovav and Beth Levin. 2010.
\newblock Reflections on manner/result complementarity.
\newblock \emph{Syntax, lexical semantics, and event structure}, pages 21--38.

\bibitem[{Ide et~al.(2008)Ide, Baker, Fellbaum, Fillmore, and
  Passonneau}]{ide2008masc}
Nancy Ide, Collin Baker, Christiane Fellbaum, Charles Fillmore, and Rebecca
  Passonneau. 2008.
\newblock Masc: The manually annotated sub-corpus of american english.
\newblock In \emph{6th International Conference on Language Resources and
  Evaluation, LREC 2008}, pages 2455--2460. European Language Resources
  Association (ELRA).

\bibitem[{Kipper et~al.(2008)Kipper, Korhonen, Ryant, and
  Palmer}]{kipper2008large}
Karin Kipper, Anna Korhonen, Neville Ryant, and Martha Palmer. 2008.
\newblock A large-scale classification of english verbs.
\newblock \emph{Language Resources and Evaluation}, 42:21--40.

\bibitem[{Krifka(1992)}]{krifka1992thematic}
Manfred Krifka. 1992.
\newblock Thematic relations as links between nominal reference and temporal
  constitution.
\newblock \emph{Lexical matters}, (24):29.

\bibitem[{Levin(2008)}]{levin2008constraint}
Beth Levin. 2008.
\newblock A constraint on verb meanings: Manner/result complementarity.
\newblock \emph{Cognitive Science Department Colloqium Series, Brown
  University, Providence, RI, March}, 17:2008.

\bibitem[{Levin and Hovav(1991)}]{levin1991wiping}
Beth Levin and Malka~Rappaport Hovav. 1991.
\newblock Wiping the slate clean: A lexical semantic exploration.
\newblock \emph{cognition}, 41(1-3):123--151.

\bibitem[{Liu et~al.(2019)Liu, Ott, Goyal, Du, Joshi, Chen, Levy, Lewis,
  Zettlemoyer, and Stoyanov}]{Liu2019RoBERTaAR}
Yinhan Liu, Myle Ott, Naman Goyal, Jingfei Du, Mandar Joshi, Danqi Chen, Omer
  Levy, Mike Lewis, Luke Zettlemoyer, and Veselin Stoyanov. 2019.
\newblock \href {https://api.semanticscholar.org/CorpusID:198953378} {Roberta:
  A robustly optimized bert pretraining approach}.
\newblock \emph{ArXiv}, abs/1907.11692.

\bibitem[{Metheniti et~al.(2022)Metheniti, Van De~Cruys, and
  Hathout}]{metheniti2022time}
Eleni Metheniti, Tim Van De~Cruys, and Nabil Hathout. 2022.
\newblock About time: Do transformers learn temporal verbal aspect?
\newblock In \emph{12th Workshop on Cognitive Modeling and Computational
  Linguistics (CMCL 2022)}, pages 88--101. ACL: Association for Computational
  Linguistic.

\bibitem[{Naigles and Hoff-Ginsberg(1998)}]{naigles1998some}
Letitia~R Naigles and Erika Hoff-Ginsberg. 1998.
\newblock Why are some verbs learned before other verbs? effects of input
  frequency and structure on children's early verb use.
\newblock \emph{Journal of child language}, 25(1):95--120.

\bibitem[{Scarborough(1990)}]{scarborough1990index}
Hollis~S Scarborough. 1990.
\newblock Index of productive syntax.
\newblock \emph{Applied psycholinguistics}, 11(1):1--22.

\bibitem[{Sennrich(2015)}]{sennrich2015neural}
Rico Sennrich. 2015.
\newblock Neural machine translation of rare words with subword units.
\newblock \emph{arXiv preprint arXiv:1508.07909}.

\bibitem[{Verhage et~al.(2020)Verhage, Schuengel, Duschinsky, van IJzendoorn,
  Fearon, Madigan, Roisman, Bakermans-Kranenburg, and
  Oosterman}]{verhage2020collaboration}
Marije~L Verhage, Carlo Schuengel, Robbie Duschinsky, Marinus~H van IJzendoorn,
  RM~Pasco Fearon, Sheri Madigan, Glenn~I Roisman, Marian~J
  Bakermans-Kranenburg, and Mirjam Oosterman. 2020.
\newblock The collaboration on attachment transmission synthesis (cats): A move
  to the level of individual-participant-data meta-analysis.
\newblock \emph{Current Directions in Psychological Science}, 29(2):199--206.

\bibitem[{Weismer et~al.(2013)Weismer, Venker, Evans, and
  Moyle}]{weismer2013fast}
Susan~Ellis Weismer, Courtney~E Venker, Julia~L Evans, and Maura~Jones Moyle.
  2013.
\newblock Fast mapping in late-talking toddlers.
\newblock \emph{Applied Psycholinguistics}, 34(1):69--89.

\bibitem[{Zhang et~al.(2023)Zhang, Li, Ma, Zhou, and Zou}]{zhang2023llmaaa}
Ruoyu Zhang, Yanzeng Li, Yongliang Ma, Ming Zhou, and Lei Zou. 2023.
\newblock Llmaaa: Making large language models as active annotators.
\newblock \emph{arXiv preprint arXiv:2310.19596}.

\end{thebibliography}

\appendix
\renewcommand{\thesubsection}{\Alph{subsection}}

\section*{Appendix}
\subsection{Instructions to Expert Annotator and Annotation Tool}\label{sec:appendixA}
\begin{figure}[h]
    \centering
    \includegraphics[width=1.0\linewidth]{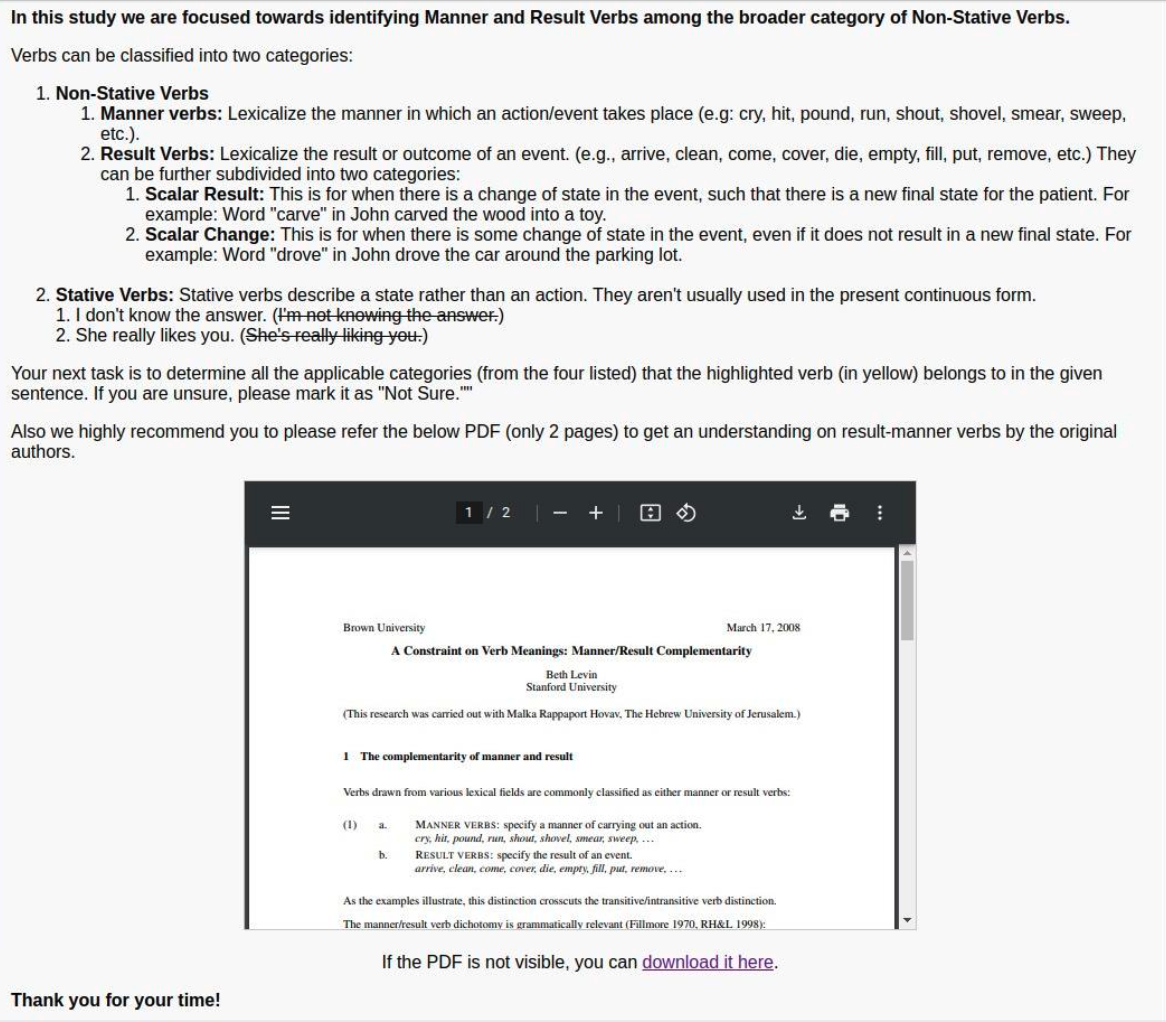}
    \caption{Annotation Screen for Expert Human Annotator.}
    \label{fig:hanr}
\end{figure}

The instructions that were provided to the expert human annotator before starting the annotation process is shown in Figure  \ref{fig:expert_guides} and the sample annotation screen is provided in Figure \ref{fig:hanr}. The users were provided with clear definition taken from \cite{hovav2010reflections, levin2008constraint} paper.

 \begin{figure}[ht!p]
\centering
\begin{mdframed}
\tiny
\noindent
\textbf{Identifying Manner and Result Verbs in Non-Stative Verbs}\\
\textbf{Definition:} Verbs can be classified into two categories: Non-Stative Verbs and Stative Verbs.\\[0.5em]

\textbf{1. Non-Stative Verbs}\\
\quad \textbf{1.1 Manner Verbs:} These verbs lexicalize the manner in which an action/event takes place. \textit{Examples:} cry, hit, pound, run, shout, shovel, smear, sweep, etc.\\[0.5em]
\quad \textbf{1.2 Result Verbs:} These verbs lexicalize the result or outcome of an event. \textit{Examples:} arrive, clean, come, cover, die, empty, fill, put, remove, etc.\\[0.5em]
\quad \textbf{1.2.1 Scalar Result:} Describes a change of state in the event, leading to a new final state. \textit{Example:} ``John \textbf{carved} the wood into a toy.''\\[0.5em]
\quad \textbf{1.2.2 Scalar Change:} Indicates some change of state in the event, even if it does not result in a new final state. \textit{Example:} ``John \textbf{drove} the car around the parking lot.''\\[1em]
\textbf{2. Stative Verbs}\\
\quad Stative verbs describe a state rather than an action. They are not typically used in the present continuous form.\\[0.5em]
\quad \textit{Examples:}\\
\quad ``I don't know the answer.'' (*I'm not knowing the answer.*) (Ungrammatical)\\
\quad ``She really likes you.'' (*She's really liking you.*) (Ungrammatical)\\[1em]

\textbf{Annotation Task:}\\
Your next task is to determine all the applicable categories (from the four listed) that the highlighted verb (in yellow) belongs to in the given sentence. If unsure, mark it as ``Not Sure.''\\[1em]

\textbf{Reference Material:}\\
For further understanding, refer to the below PDF (only 2 pages) for insights on manner-result verbs by the original authors.\\[0.5em]

\end{mdframed}
\caption{Guidelines for Identifying Manner and Result Verbs in Non-Stative Verbs}
\label{fig:expert_guides}
\end{figure}

\begin{figure}[h]  
    \centering
    \includegraphics[width=1.0\linewidth]{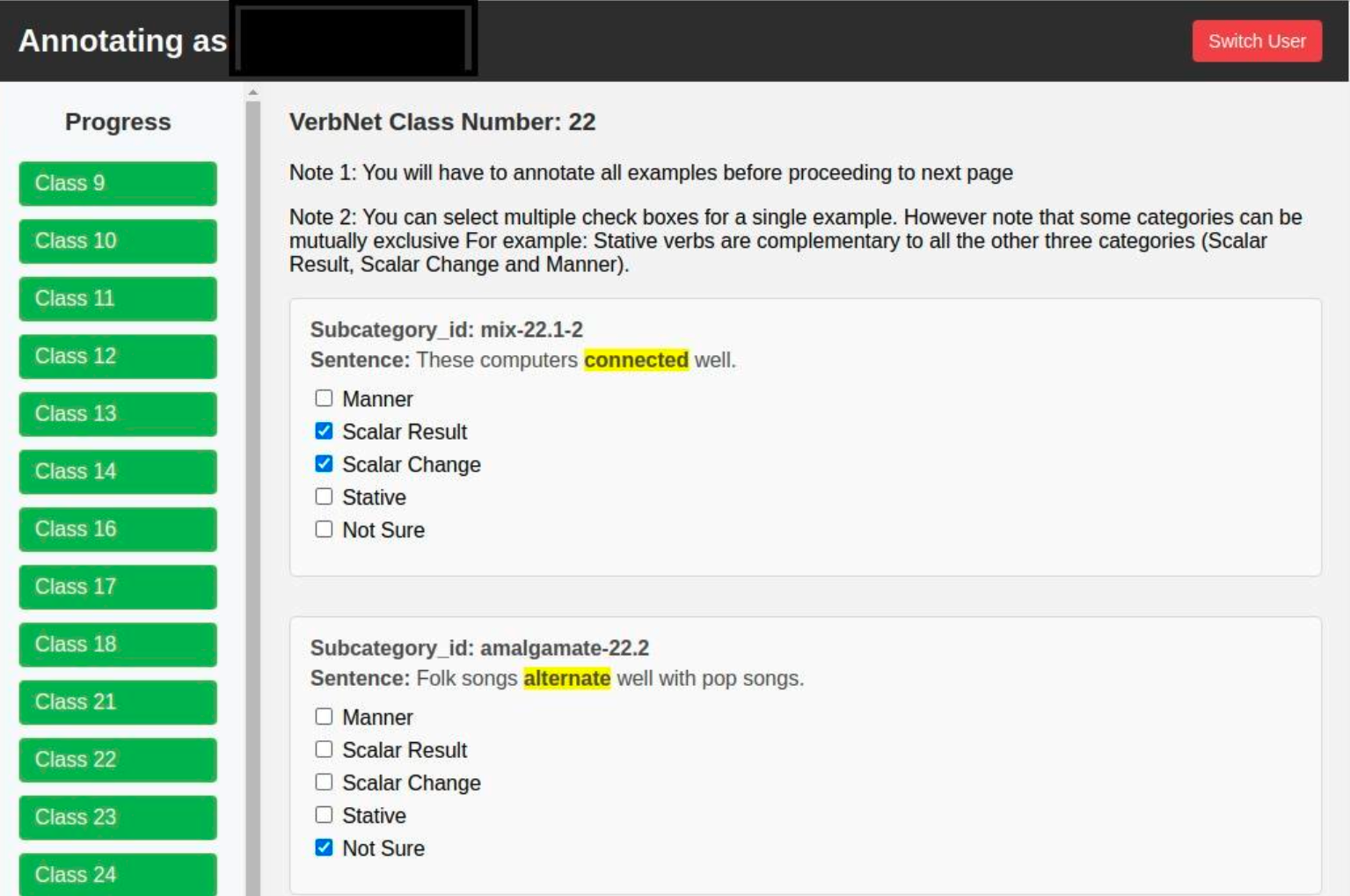}
    \caption{Sample Annotation Screen.}
    \label{fig:anoScreen}
\end{figure}

A sample annotation screen is shown in Figure \ref{fig:anoScreen}. The user can tag the sentences in multiple sessions and there were a total of 200 sentences to annotate. The VerbNet categories are shown on the left.

\subsection{LLM Prompting}\label{sec:appendixB}
Figure \ref{fig:sentenceConstruction} represents the rule for instructing LLM to focus on the sentence construction while tagging result and manner verbs.

\begin{figure}[ht!p]
\centering
\begin{mdframed}
\tiny
\noindent
\textbf{Manner Verbs}\\
\textbf{Definition:} These verbs encode the *how* of an action, focusing on the method or process by which an action is performed rather than its outcome.\\[0.5em]
\textbf{Syntactic Diagnostic 1: Unspecified Objects}\\
Manner verbs frequently occur with unspecified or non-subcategorized objects in nonmodal, nonhabitual sentences.\\
\textit{Example:} ``Anna wept all day.'' (Acceptable)\\[0.5em]
\textbf{Syntactic Diagnostic 2: Causative/Inchoative Alternation}\\
Manner verbs do not participate in the causative/inchoative alternation.\\
\textit{Example:}  
\quad Causative: ``John wiped the table.''\\
\quad Inchoative: *``The table wiped.''* (Ungrammatical)\\[0.5em]
\textbf{Usage:}\\
\quad ``She \textbf{scribbled} on the notebook.'' (Focus on the method of writing)\\[1em]

\textbf{Result Verbs}\\
\textbf{Definition:} These verbs encode the *outcome* or resultant state that follows from an action.\\[0.5em]
\textbf{Syntactic Diagnostic 1: Specified Objects}\\
Result verbs typically do not occur with unspecified or non-subcategorized objects. They require a direct object that undergoes a change.\\[0.5em]
\textbf{Syntactic Diagnostic 2: Causative/Inchoative Alternation}\\
Result verbs readily participate in the causative/inchoative alternation, appearing both in causative constructions (with an explicit external agent) and in inchoative constructions (where the change occurs spontaneously).\\
\textit{Examples:}\\
\quad Causative: ``The child broke the vase.'' (Agent causes the change)\\
\quad Inchoative: ``The vase broke.'' (The change occurs without an explicit agent)\\[0.5em]
\textbf{Usage:}\\
\quad ``He \textbf{melted} the ice.'' (Focus on the resulting state)
\end{mdframed}
\caption{Manner vs. Result Verb Sentence Construction Prompt}
\label{fig:sentenceConstruction}
\end{figure}

\subsection{Qualitative Analysis}\label{sec:appendixC}

Here we illustrate some qualitative cases where, given a sentence as input, we checked the categorization returned by the two models. Both models could identify the distinct nuances between manner and result verbs in most cases. For example, in the sentence "\emph{She sponged the bottle well}" both models correctly classified the verb "\emph{sponged}" as a manner verb, while in the sentence "\emph{She cleaned the bottle well}", both models accurately classified the verb "cleaned" as a result verb. This demonstrates that, irrespective of context, the models developed an understanding of the verb root to distinguish between result and manner connotations.

Additionally, to highlight the capability of the models in distinguishing stative and non-stative verbs, we checked a few sentences. In the sentence "\emph{The mother ran to the market and bought her child a gift, because she loves her a lot}", both models accurately identified the categories of the verbs "\emph{ran}", "\emph{bought}", and "\emph{loves}" as manner, result, and stative, respectively. However, when given the sentence, \\
"\emph{The president learned of a coup plot that might endanger his life}", model 2 incorrectly classified the verb "\emph{endanger}" as stative, while model 1 accurately identified the verb as result.

\end{document}